\documentclass[journal]{IEEEtran}
\usepackage{amsmath,amsfonts}
\usepackage{algorithmic}
\usepackage{algorithm}
\usepackage{amsmath}
\usepackage{array}
\usepackage[caption=false,font=normalsize,labelfont=sf,textfont=sf]{subfig}
\usepackage{tabularx}
\usepackage{textcomp}
\usepackage{stfloats}
\usepackage{url}
\usepackage{verbatim}
\usepackage{graphicx}
\usepackage{cite}
\usepackage{hyperref}
\usepackage{multirow}
\usepackage{tabularx}
\usepackage{makecell}
\usepackage{booktabs}
\usepackage{adjustbox}
\begin{document}

\title{Brain-Inspired Graph Multi-Agent Systems for LLM Reasoning}

\author{
  \textbf{Guangfu Hao\textsuperscript{1,2,†}}, 
  \textbf{Yuming Dai\textsuperscript{3,†}}, 
  \textbf{Xianzhe Qin\textsuperscript{4}}, 
  \textbf{Shan Yu\textsuperscript{1,2,5,*}\thanks{†These authors contributed equally to this work. *Corresponding authors: \href{mailto:shan.yu@nlpr.ia.ac.cn}{shan.yu@nlpr.ia.ac.cn}}}

  \textsuperscript{1}Laboratory of Brain Atlas and Brain-inspired Intelligence, Institute of Automation,\\Chinese Academy of Sciences (CASIA)\\
  \textsuperscript{2}School of Artificial Intelligence, University of Chinese Academy of Sciences (UCAS)\\
  \textsuperscript{3}University of the Chinese Academy of Sciences (UCAS)\\
  \textsuperscript{4}College of Software (CS), Taiyuan University of Technology\\
  \textsuperscript{5}School of Future Technology, University of Chinese Academy of Sciences (UCAS)\\
}

\maketitle

\begin{abstract}
Large Language Models (LLMs) have demonstrated remarkable capabilities across a wide range of language tasks, yet complex multi-step reasoning remains a fundamental challenge. While Large Reasoning Models (LRMs) equipped with extended chain-of-thought mechanisms demonstrate improved performance over standard LLMs, both model types still suffer from accuracy collapse on sufficiently complex tasks~\cite{zheng2025curse, he2025can}, suggesting that scaling model-level reasoning alone is insufficient. Inspired by the global workspace theory of human cognition, we propose \textbf{Brain-Inspired Graph Multi-Agent Systems (BIGMAS)}, in which specialized LLM agents are organized as nodes in a dynamically constructed directed graph and coordinate exclusively through a centralized shared workspace. A problem-adaptive GraphDesigner constructs task-specific agent topologies, while a global Orchestrator leverages the complete shared state for routing decisions, overcoming the local-view bottleneck of reactive approaches. Experiments on Game24, Six Fives, and Tower of London across six frontier LLMs demonstrate that BIGMAS consistently improves reasoning performance for both standard LLMs and LRMs, outperforming existing multi-agent baselines including ReAct and Tree of Thoughts, showing that multi-agent architectural design provides complementary gains orthogonal to model-level reasoning enhancements.
\end{abstract}

\begin{IEEEkeywords}
Multi-agent systems, large language models, graph-based reasoning, shared workspace coordination, dynamic graph design, complex problem solving
\end{IEEEkeywords}

\section{Introduction}

Recent generations of frontier models have introduced Large Reasoning Models (LRMs) 
such as OpenAI o1/o3~\cite{jaech2024openai, openai2025o3}, 
DeepSeek-R1~\cite{guo2025deepseek}, Claude Sonnet Thinking~\cite{anthropic2025claude45sonnet}, 
and Gemini Thinking~\cite{comanici2025gemini}, characterized by extended 
chain-of-thought (CoT)~\cite{wei2022chain} mechanisms with self-reflection~\cite{renze2024self}. 
While these models demonstrate improved performance on standard reasoning 
benchmarks~\cite{ferrag2025llm, cao2025toward}, recent systematic investigations 
reveal that both standard LLMs and LRMs suffer from accuracy collapse beyond 
certain problem complexity thresholds~\cite{shojaee2025illusion}. Moreover, 
providing explicit solution algorithms to LRMs does not alleviate this collapse, 
suggesting the limitation lies not merely in solution discovery but in consistent logical execution—pointing to a fundamental bottleneck that model-level scaling alone cannot resolve.

This observation motivates a complementary perspective: rather than continuing to 
scale individual model reasoning, can we design \emph{multi-agent architectures} 
that distribute cognitive load across specialized components and externalize 
intermediate reasoning state? While multi-agent LLM frameworks have emerged as 
a promising direction~\cite{hong2023metagpt, liu2023agentbench, wang2024survey,xi2025rise}, existing approaches 
share a critical structural limitation: agent communication is either point-to-point 
or encoded in fixed, pre-specified topologies, leaving global task state fragmented across individual agents, preventing both full-state visibility and dynamic adaptation of their collaborative organization~\cite{dong2024large,jimenez2025multi} to the specific demands of each problem.

A deeper inspiration comes from the organizational principles of the human brain. 
Under \emph{Global Workspace Theory} (GWT)~\cite{baars1988global, dehaene2011experimental}, 
flexible cognition emerges from the dynamic formation of coalitions among distributed 
specialized processors, coordinated through a shared central workspace. Critically, 
both the \emph{composition} of the coalition and the \emph{topology} of their 
interactions are determined by task demands rather than hardwired in advance. 
Translating these principles into a multi-agent system suggests an architecture 
in which: (i) a meta-level agent \emph{designs} a task-specific graph of specialized 
agents per problem; (ii) the resulting graph is adaptive—different problems yield 
different node compositions and topologies; and (iii) all agents coordinate through 
a shared workspace that maintains a globally consistent view of task state.

In this paper, we present \textbf{Brain-Inspired Graph Multi-Agent Systems (BIGMAS)}, 
a novel framework that instantiates these principles for LLM reasoning. 
A \textit{GraphDesigner} agent autonomously constructs a task-specific directed 
agent graph together with a shared workspace schema for each problem. 
All agent nodes interact exclusively through the centralized shared workspace, 
ensuring every intermediate result is globally visible. A global \textit{Orchestrator} 
observes the complete workspace state and full execution history at each routing step, 
eliminating the local-view bottleneck of reactive paradigms. Agent outputs are 
validated against workspace constraints, with a self-correction loop that resolves 
errors without aborting execution.

We evaluate BIGMAS on three reasoning benchmarks—Game24, Six Fives, and Tower of London—across six frontier LLMs (DeepSeek, Claude, GPT, and Gemini). 
BIGMAS consistently improves performance for both standard LLMs and LRMs. 
The gains are largest precisely where individual models struggle most, suggesting 
that multi-agent coordination provides a structural remedy for reasoning collapse 
that is orthogonal to model-level capability. BIGMAS also outperforms established multi-agent baselines---ReAct and 
Tree of Thoughts---across all three tasks, confirming that the gains stem 
from the synergy of adaptive topology and global workspace coordination 
rather than from more extensive search alone. Our key contributions are as follows:

\begin{itemize}
    \item \textbf{Brain-Inspired Dynamic Graph Architecture.} We propose BIGMAS, 
    grounded in global workspace theory. A GraphDesigner agent autonomously constructs 
    a task-specific directed agent graph with a shared workspace schema per problem, 
    replacing static topologies with problem-adaptive structures.

    \item \textbf{Global-State Orchestration with Robust Execution.} A global 
Orchestrator makes routing decisions based on the complete shared workspace 
state and full execution history, overcoming the local-view bottleneck of 
reactive paradigms. A self-correction loop with multi-strategy fallback 
parsing ensures execution integrity without aborting on transient node failures.

    \item \textbf{Comprehensive Empirical Validation.} Experiments across three 
    reasoning benchmarks and six frontier LLMs demonstrate consistent and substantial 
    improvements over single-model baselines for both standard LLMs and LRMs, 
    establishing complementary gains orthogonal to model-level enhancements.
\end{itemize}

\begin{figure*}[ht]
    \centering
    \includegraphics[width=0.8\linewidth]{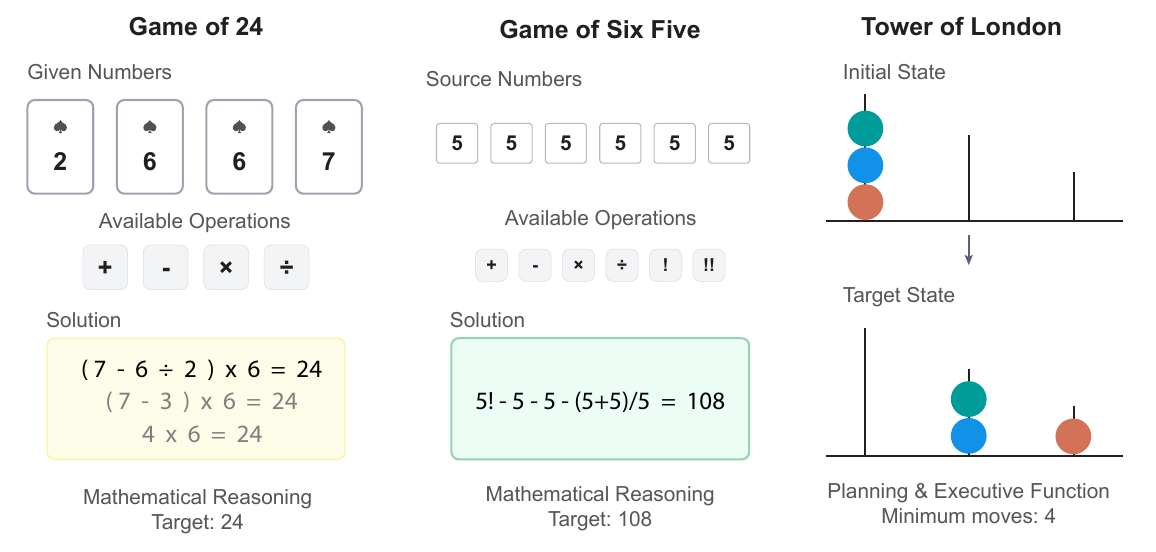}
    \caption{\textbf{Three cognitive reasoning tasks used in evaluation.} Left: Tower of London task requires planning optimal moves to reach target configuration. Middle: Six Fives requires constructing arithmetic expressions using exactly six 5s to reach a target value. Right: Game24 demands mathematical reasoning to combine four numbers reaching target value 24.}
    \label{fig:1}
\end{figure*}

\section{Related Work}

\subsection{LLM Reasoning and Multi-Agent Frameworks}

Improvements in LLM reasoning have followed two broad trajectories. The first 
scales inference-time computation through prompting strategies such as 
Chain-of-Thought~\cite{wei2022chain} and self-consistency~\cite{wang2022self}, 
decomposition strategies such as least-to-most prompting~\cite{zhou2022least} 
and Program of Thoughts~\cite{chen2022program}, or iterative self-refinement 
approaches~\cite{madaan2023selfrefine}, or through reinforcement learning from 
verifiable rewards~\cite{guo2025deepseek}, 
yielding LRMs that generate extended reasoning traces before producing answers. 
Shojaee et al.~\cite{shojaee2025illusion} provide a systematic characterization 
of LRM limitations: accuracy collapses beyond problem-specific complexity thresholds, 
reasoning effort counterintuitively decreases near the collapse point, and the 
collapse persists even when explicit solution algorithms are provided. 
These findings establish that model-level scaling hits a fundamental ceiling 
on sufficiently complex tasks.

The second trajectory develops multi-agent frameworks that distribute reasoning 
across specialized components. ReAct~\cite{yao2022react} interleaves reasoning 
and action in a reactive loop; Reflexion~\cite{shinn2023reflexion} augments this 
with verbal self-reflection; Tree of Thoughts~\cite{yao2023tree} broadens 
exploration via tree search. Graph of Thoughts~\cite{besta2024got} extends this further by modeling LLM-generated information as an arbitrary graph where units of reasoning are vertices and dependencies are edges, enabling feedback loops and thought merging. Buffer of Thoughts~\cite{yang2024buffer} proposes a meta-buffer of reusable thought-templates to improve accuracy, efficiency and robustness across diverse reasoning tasks. These approaches improve on single-pass inference 
but remain limited by incremental, partial-information decision-making or by 
operating within a single model's context. Role-specialized frameworks such as MetaGPT~\cite{hong2023metagpt} introduce structured collaboration, while planning-execution decoupling via DAG structures—LLMCompiler~\cite{kim2024llm} and the planner-centric paradigm of Wei et al.~\cite{wei2025beyond}—demonstrates the advantage of global upfront planning over reactive approaches. Multi-agent conversation frameworks such as AutoGen~\cite{wu2024autogen} and role-playing frameworks such as CAMEL~\cite{li2023camel} enable flexible agent orchestration through programmable interaction patterns. Multiagent debate~\cite{du2024improving} demonstrates that having multiple model instances propose and critique each other's reasoning can substantially improve factual accuracy and mathematical reasoning. Dynamic agent team selection approaches such as DyLAN~\cite{liu2023dynamic} and graph-based optimization frameworks such as GPTSwarm~\cite{zhuge2024gptswarm} show that varying agent composition and communication topology according to the task improves over fixed-structure alternatives. 
Despite this progress, all existing frameworks share the limitation that 
coordination graphs are fixed and agent state is not globally shared, 
preventing dynamic adaptation to problem structure. BIGMAS addresses this 
gap by combining per-problem graph construction with a centralized shared 
workspace, connecting both trajectories into a unified architecture.

\subsection{Neuroscience Foundations: Dynamic Networks and Global Workspace}

Systems neuroscience has established that the brain does not use a fixed 
processing pipeline. Large-scale functional networks reconfigure dynamically 
according to task demands~\cite{sporns2010networks, bullmore2009complex}, 
with prefrontal hub regions orchestrating the selective recruitment of 
specialized areas~\cite{cole2013multi}. Task complexity modulates the 
degree of cross-network coordination required: routine tasks engage relatively 
isolated modules, while complex or novel problems necessitate broader 
functional coalitions~\cite{bassett2011dynamic}. Recent neuroimaging studies confirm that these functional network reconfigurations are not merely structural but reflect training-induced plasticity~\cite{finc2020dynamic}, further supporting the view that task-adaptive topology is a core computational principle of flexible cognition.

Global Workspace Theory (GWT)~\cite{baars1988global, dehaene2011experimental} 
formalizes this observation. Under GWT, flexible cognition arises when 
specialized processors form a task-driven coalition and communicate through 
a central workspace that broadcasts information globally to all recruited modules. 
The three core properties of GWT—processor specialization, dynamic coalition 
formation, and global broadcast—map directly onto design principles for 
multi-agent systems: distinct agent roles, per-problem graph construction, 
and a centralized shared workspace visible to all agents. Existing LLM 
frameworks instantiate specialization but not the other two properties; 
BIGMAS is the first to operationalize all three within a unified architecture, 
grounding its design in an empirically supported theory of flexible cognition.

\begin{figure*}[ht]
    \centering
    \includegraphics[width=\linewidth]{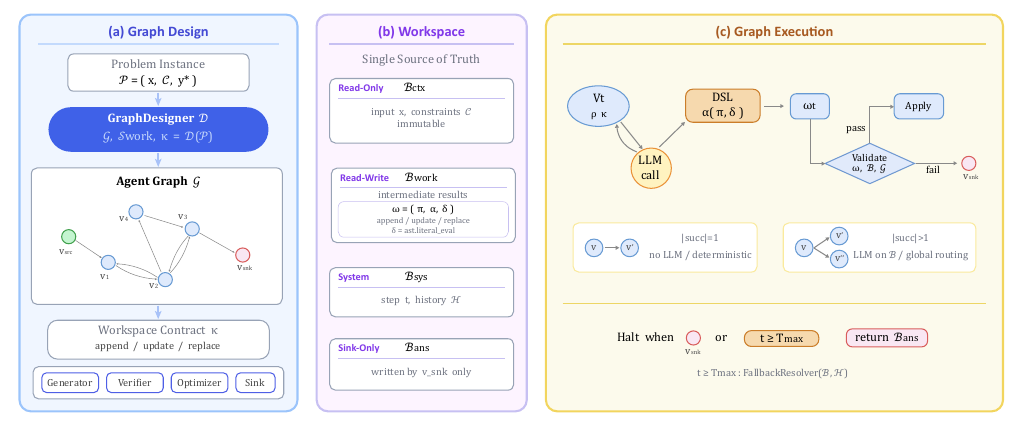}
    \caption{\textbf{Overview of the BIGMAS framework.} 
    (a)~\textbf{Graph Design}: A GraphDesigner agent $\mathcal{D}$ analyzes 
    the problem instance $\mathcal{P} = (x, \mathcal{C}, y^*)$ and produces 
    a task-specific directed agent graph $\mathcal{G}$ together with a 
    Workspace contract $\kappa$. 
    (b)~\textbf{Workspace} $\mathcal{B}$: A centralized shared workspace 
    partitioned into read-only context $\mathcal{B}_{\text{ctx}}$, 
    read-write working area $\mathcal{B}_{\text{work}}$, system metadata 
    $\mathcal{B}_{\text{sys}}$, and sink-only answer store $\mathcal{B}_{\text{ans}}$; 
    all agent nodes interact exclusively through $\mathcal{B}$. 
    (c)~\textbf{Graph Execution}: Each active node $v_t$ produces a 
    structured write instruction $\omega_t = (\pi, \alpha, \delta)$ via an 
    LLM call; the instruction is validated against $\mathcal{B}$ and $\mathcal{G}$, 
    with a self-correction loop on failure. A global Orchestrator routes 
    execution based on the complete workspace state; the system halts 
    when the sink node $v_{\text{snk}}$ is reached or the step budget 
    $T_{\max}$ is exhausted.}
    \label{fig:overview}
\end{figure*}

\subsection{Reasoning Benchmarks and Evaluation}

Widely used benchmarks such as GSM8K~\cite{cobbe2021training}, 
MATH~\cite{hendrycks2021measuring}, and HumanEval~\cite{chen2021evaluating} 
have driven LLM progress but present two well-known limitations for systematic 
analysis: susceptibility to data contamination in frontier model training 
corpora~\cite{shojaee2025illusion}, and fixed problem complexity that 
precludes controlled study of performance scaling. Evaluation is also 
typically restricted to final answer accuracy, obscuring intermediate 
reasoning quality.

Controllable puzzle environments overcome these limitations by enabling 
precise complexity manipulation and simulator-based step-level 
verification~\cite{shojaee2025illusion, valmeekam2023planning}. 
The three benchmarks in this work—Game24~\cite{yao2023tree}, Six Fives, 
and Tower of London~\cite{shallice1982specific}—provide 
contamination-resistant, verifiable tasks spanning arithmetic reasoning, 
combinatorial search, and sequential planning. 
Unlike tool-augmented evaluations such as ToolBench~\cite{qin2023toolllm} 
and StableToolBench~\cite{guo2024stabletoolbench}, where task success 
conflates reasoning quality with tool reliability, these benchmarks isolate 
the reasoning contribution directly, enabling a clean measurement of 
architectural gains across both standard LLMs and LRMs.

\section{Method}
\label{sec:methodology}

\subsection{Problem Formulation}

We consider a class of combinatorial reasoning tasks that require multi-step 
logical deduction, constraint satisfaction, or sequential planning. Formally, 
a reasoning problem instance is defined as a tuple $\mathcal{P} = (x, \mathcal{C}, y^*)$, 
where $x$ denotes the problem input (e.g., a set of numbers, an initial board 
configuration, or a hidden code), $\mathcal{C}$ denotes the set of task-specific 
constraints (e.g., operator restrictions, move legality rules, or capacity limits), 
and $y^*$ denotes the target output (e.g., a valid expression, a move sequence, 
or a sequence of guesses). The goal is to produce a solution $\hat{y}$ that 
satisfies all constraints in $\mathcal{C}$ and achieves the target $y^*$.

The three tasks considered in this work instantiate this formulation as follows.

\textbf{Game24.} Given four integers $\{n_1, n_2, n_3, n_4\} \subset [1, 13]$, 
find an arithmetic expression using each number exactly once with operators 
$\{+, -, \times, \div\}$ and parentheses such that the result equals 24. 
Formally, find $\hat{y} = f(n_1, n_2, n_3, n_4)$ such that $\text{eval}(\hat{y}) = 24$.

\textbf{Six Fives.} Given a target integer $t$, construct an arithmetic expression 
using exactly six instances of the digit 5—with concatenation (e.g., $55$, $555$) 
permitted—and operators $\{+, -, \times, \div, !, !!\}$ such that the expression 
evaluates to $t$. The constraint set $\mathcal{C}$ requires that the count of 
digit 5 used equals exactly six.

\textbf{Tower of London.} Given three pegs with capacities $(3, 2, 1)$ and three 
colored beads $\{r, g, b\}$, find a minimum-length sequence of moves 
$\hat{y} = (m_1, m_2, \ldots, m_k)$ that transforms an initial peg configuration 
$s_0$ into a goal configuration $s^*$. Each move $m_i$ transfers the topmost bead 
of one peg to another, subject to peg capacity constraints.

\subsection{BIGMAS Architecture Overview}

BIGMAS solves a reasoning problem $\mathcal{P}$ through a three-phase pipeline: 
\textit{graph design}, \textit{graph execution}, and \textit{answer extraction}. 
The system maintains a centralized \textit{Workspace} $\mathcal{B}$ as the single 
source of truth throughout execution. All agent nodes interact exclusively with 
$\mathcal{B}$; no point-to-point communication between nodes occurs. The complete framework is illustrated in Figure~\ref{fig:overview}.

The Workspace is structured as a four-partition data object:
\begin{equation}
    \mathcal{B} = \left( \mathcal{B}_{\text{ctx}},\; 
                         \mathcal{B}_{\text{work}},\; 
                         \mathcal{B}_{\text{sys}},\; 
                         \mathcal{B}_{\text{ans}} \right)
\end{equation}
where $\mathcal{B}_{\text{ctx}}$ stores the read-only problem context (input $x$ 
and constraints $\mathcal{C}$); $\mathcal{B}_{\text{work}}$ is the read-write 
working area for intermediate results; $\mathcal{B}_{\text{sys}}$ records system 
metadata including execution step count and routing history; and 
$\mathcal{B}_{\text{ans}}$ holds the final answer written by the terminal node.

\subsection{Phase 1: Problem-Adaptive Graph Design}

Given a problem instance $\mathcal{P}$, a \textit{GraphDesigner} agent $\mathcal{D}$ 
analyzes the problem structure and produces a task-specific directed agent graph 
together with a Workspace working-area schema:
\begin{equation}
    \left( \mathcal{G}, \mathcal{S}_{\text{work}}, \kappa \right) 
    = \mathcal{D}(\mathcal{P})
\end{equation}
where $\mathcal{G} = (\mathcal{V}, \mathcal{E}, v_{\text{src}}, v_{\text{snk}})$ 
is a directed graph with node set $\mathcal{V}$, edge set $\mathcal{E}$, 
a designated source node $v_{\text{src}}$, and a sink node $v_{\text{snk}}$; 
$\mathcal{S}_{\text{work}}$ is the initial template for $\mathcal{B}_{\text{work}}$; 
and $\kappa$ is a natural-language \textit{Workspace contract} specifying each 
node's read and write responsibilities.

Each node $v_i \in \mathcal{V}$ is characterized by a role descriptor 
$\rho_i$ (e.g., \textit{expression generator}, \textit{validator}, 
\textit{strategy updater}) and a set of Workspace interaction 
permissions derived from $\kappa$. The graph $\mathcal{G}$ may contain 
cycles to support iterative refinement. Different problem instances 
yield structurally distinct graphs, reflecting the dynamic coalition 
formation principle of Global Workspace Theory.

\begin{algorithm}[t]
\caption{BIGMAS Execution}
\label{alg:bigmas}
\begin{algorithmic}[1]
\REQUIRE Problem $\mathcal{P}$, step budget $T_{\max}$, 
         correction budget $R$
\ENSURE  Final answer $\mathcal{B}_{\text{ans}}$

\STATE $(\mathcal{G}, \mathcal{S}_{\text{work}}, \kappa) 
       \leftarrow \mathcal{D}(\mathcal{P})$
       \hfill\COMMENT{Graph design}
\STATE Initialize Workspace $\mathcal{B}$ with $\mathcal{B}_{\text{ctx}} \leftarrow \mathcal{P}$, 
       $\mathcal{B}_{\text{work}} \leftarrow \mathcal{S}_{\text{work}}$
\STATE $v \leftarrow v_{\text{src}}$;\quad 
       $t \leftarrow 0$;\quad 
       $\mathcal{H} \leftarrow \{\}$
\WHILE{$v \neq v_{\text{snk}}$ \AND $t < T_{\max}$}
    \STATE $\omega \leftarrow v(\mathcal{B}, \rho_v, \kappa)$
           \hfill\COMMENT{Node execution}
    \FOR{$r = 0$ \TO $R-1$}
        \STATE $(\sigma, \epsilon) \leftarrow 
               \text{Validate}(\omega, \mathcal{B}, \mathcal{G})$
        \IF{$\sigma = \texttt{pass}$}
            \STATE \textbf{break}
        \ENDIF
        \STATE $\omega \leftarrow v(\mathcal{B}, \rho_v, \kappa, \epsilon)$
               \hfill\COMMENT{Self-correction}
    \ENDFOR
    \IF{$\sigma = \texttt{pass}$}
        \STATE $\mathcal{B} \leftarrow \text{Apply}(\omega, \mathcal{B})$
        \STATE $\mathcal{H} \leftarrow \mathcal{H} \cup \{(v, \omega)\}$
        \STATE $v \leftarrow \mathcal{O}(\mathcal{B}, \mathcal{H}, 
               \text{succ}(v), \mathcal{G})$
               \hfill\COMMENT{Orchestrator routing}
    \ELSE
        \STATE $v \leftarrow v_{\text{snk}}$
               \hfill\COMMENT{Route to sink on repeated failure}
    \ENDIF
    \STATE $t \leftarrow t + 1$
\ENDWHILE
\IF{$t \geq T_{\max}$}
    \STATE $\mathcal{B}_{\text{ans}} \leftarrow 
           \text{FallbackResolver}(\mathcal{B}, \mathcal{H})$
\ENDIF
\RETURN $\mathcal{B}_{\text{ans}}$
\end{algorithmic}
\end{algorithm}

\subsection{Phase 2: Graph Execution}

Execution proceeds as an iterative read-execute-write loop driven by 
a \textit{GraphExecutor}. Let $v_t$ denote the active node at step $t$, 
and let $\mathcal{B}^{(t)}$ denote the Workspace state at the beginning 
of step $t$. The executor maintains an execution history 
$\mathcal{H}^{(t)} = \{(v_\tau, \omega_\tau)\}_{\tau=1}^{t-1}$, 
where $\omega_\tau$ records the write operation performed at step $\tau$.

\subsubsection{Node Execution and Write Protocol}

Each node $v_t$ receives $\mathcal{B}^{(t)}$, its role descriptor $\rho_t$, 
and contract $\kappa$ as context, and produces a structured write instruction:

\begin{equation}
    \omega_t = v_t\!\left(\mathcal{B}^{(t)},\, \rho_t,\, \kappa\right)
\end{equation}

Each write instruction $\omega_t = (\pi_t, \alpha_t, \delta_t)$ specifies a target path $\pi_t$ within $\mathcal{B}_{\text{work}}$, an action 
$\alpha_t \in \{\texttt{append}, \texttt{update}, \texttt{replace}\}$, 
and a payload $\delta_t$ containing the node's output data. 
The \texttt{append} action adds an element to a list field; 
\texttt{update} merges key-value pairs into a dict field; 
and \texttt{replace} overwrites a field entirely. 
Node outputs are produced in a structured natural-language format 
that explicitly declares the target path, action type, and payload, 
which a multi-strategy parser with fallback decoding extracts into 
$\omega_t$, maximizing robustness across output format variations.

\subsubsection{Write Validation and Self-Correction}

Before applying $\omega_t$ to the Workspace, the executor performs 
a validation check:
\begin{equation}
    (\sigma_t, \epsilon_t) = \text{Validate}(\omega_t,\, \mathcal{B}^{(t)},\, \mathcal{G})
\end{equation}
where $\sigma_t \in \{\texttt{pass}, \texttt{fail}\}$ and $\epsilon_t$ 
is a structured error message. Validation enforces three conditions: 
(i) the target path $\pi_t$ exists in $\mathcal{B}_{\text{work}}$ or 
equals $\mathcal{B}_{\text{ans}}$ for the sink node; 
(ii) the action $\alpha_t$ is compatible with the type of the target field 
(\texttt{append} requires a list, \texttt{update} requires a dict); 
and (iii) the payload $\delta_t$ is non-empty.

If $\sigma_t = \texttt{fail}$, a self-correction loop re-invokes node $v_t$ 
with $\epsilon_t$ appended to its context, at most $R$ times:
\begin{equation}
    \omega_t^{(r+1)} = v_t\!\left(\mathcal{B}^{(t)},\, \rho_t,\, \kappa,\, \epsilon_t^{(r)}\right), 
    \quad r = 0, 1, \ldots, R-1
\end{equation}
If validation passes after correction, $\omega_t$ is applied and 
$\mathcal{B}^{(t)}$ is updated to $\mathcal{B}^{(t+1)}$. The Workspace 
state is never modified by a failed write, preserving integrity throughout execution.

\subsubsection{Global Orchestration and Routing}

After each successful write, a global \textit{Orchestrator} $\mathcal{O}$ 
determines the next active node. For nodes with a unique successor, routing 
is deterministic. For branching nodes with $|\text{succ}(v_t)| > 1$, 
the Orchestrator conditions on the full Workspace state and execution history:
\begin{equation}
    v_{t+1} = \mathcal{O}\!\left(\mathcal{B}^{(t+1)},\, \mathcal{H}^{(t)},\, 
    \text{succ}(v_t),\, \mathcal{G}\right)
\end{equation}
This global conditioning allows the Orchestrator to detect convergence 
(e.g., a validated solution already exists in $\mathcal{B}_{\text{work}}$), 
identify unproductive cycles, and route to fallback nodes when needed. 
To prevent non-termination, a step budget $T_{\max}$ is enforced; 
if $t > T_{\max}$, a \textit{FallbackResolver} directly extracts the 
best available answer from $\mathcal{B}^{(t)}$. The FallbackResolver is invoked when the step budget \( T_{\text{max}} \) is exhausted before the sink node is reached. It scans \( \mathcal{B}_{\text{work}} \) for any candidate answer written during execution—prioritizing entries that pass constraint verification—and writes the best available candidate to \( \mathcal{B}_{\text{ans}} \). If no valid candidate exists, it returns the most recent non-empty output as a best-effort answer. The complete execution procedure is summarized in Algorithm~\ref{alg:bigmas}.

\subsection{Experimental Setup}

\textbf{Tasks and Datasets.}
We evaluate on three reasoning benchmarks.
\textit{Game~24}: 100 instances sampled from the original 1{,}362 problems~\cite{yao2023tree} 
using stratified sampling within one standard deviation of problem difficulty.
\textit{Six~Fives}: 100 instances with integer targets drawn uniformly from the range $[1, 100]$.
\textit{Tower of London}: 100 instances sampled uniformly across optimal solution lengths 
of 1 to 8 moves.

\textbf{BIGMAS Configuration.}
The GraphDesigner is constrained to produce graphs with at most $\texttt{MAX\_NODES} = 10$ 
nodes and paths of at most $\texttt{MAX\_PATH\_LENGTH} = 15$ steps. 
All LLM calls within BIGMAS use a sampling temperature of $0.7$.

\textbf{Baselines.}
We compare against three baselines using DeepSeek-V3.2 as the backbone model.
\textit{Base LLM}: direct single-call inference with no additional scaffolding.
\textit{ReAct}~\cite{yao2022react}: a reactive reasoning-and-acting loop 
with a maximum of 10 turns and temperature $0.7$.
\textit{Tree of Thoughts}~\cite{yao2023tree}: tree-structured search 
with $\texttt{max\_rounds} = 4$, $\texttt{n\_thoughts} = 3$, and temperature $0.7$.

\textbf{Evaluation Metric.}
All methods are evaluated using accuracy (\%), defined as the percentage of instances 
for which the system produces a solution that satisfies all task constraints and 
achieves the specified target value.

\section{Results}
\label{sec:results}

\subsection{Overall Performance}

\begin{figure*}[htbp]
  \centering
  \includegraphics[width=\linewidth]{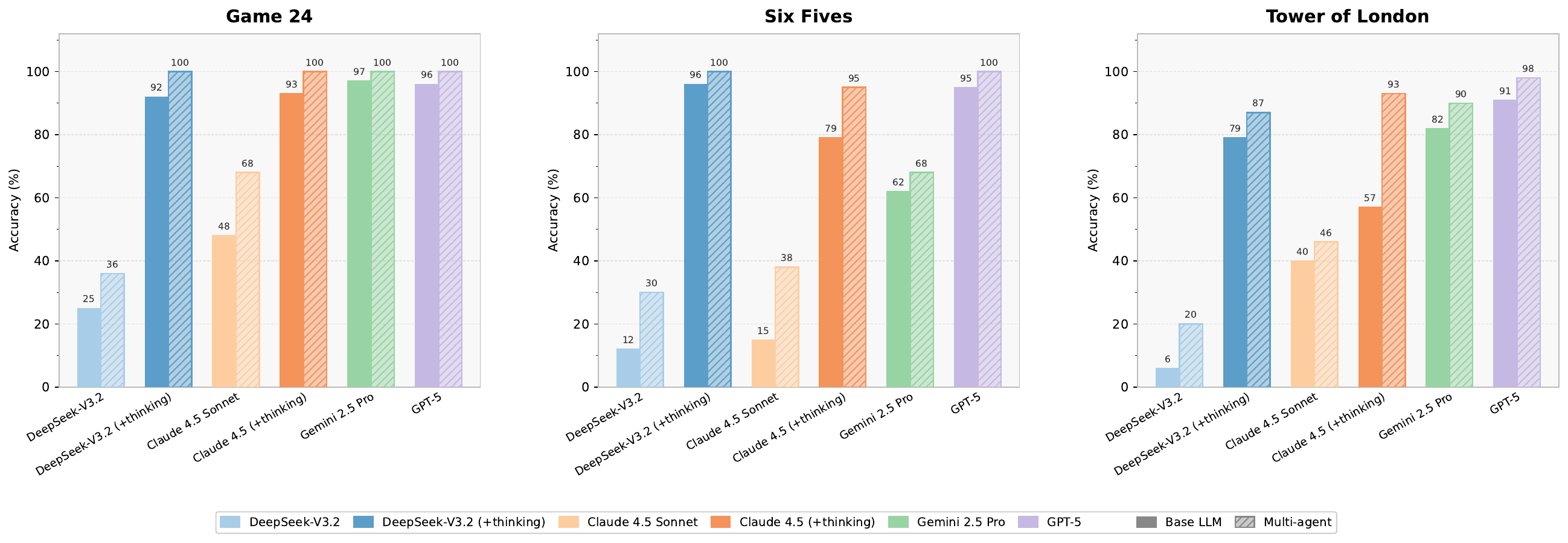}
  \caption{%
    Grouped bar chart comparing the accuracy (\%) of six LLMs on three
    reasoning benchmarks under two conditions: \textit{Base LLM} (solid bars)
    and \textit{BIGMAS} (hatched bars).
    The six models evaluated are DeepSeek-V3.2, DeepSeek-V3.2 (+thinking),
    Claude~4.5 Sonnet, Claude~4.5 (+thinking), Gemini~2.5 Pro, and GPT-5.
    The three tasks are \textbf{Game~24} (arithmetic reasoning),
    \textbf{Six~Fives} (constrained expression generation), and
    \textbf{Tower of London} (multi-step planning).
    For every model and task, BIGMAS consistently matches
    or improves upon the corresponding base-LLM accuracy, with particularly
    large gains on weaker base models (e.g., DeepSeek-V3.2 and Claude~4.5
    Sonnet) and smaller but still positive gains on already strong models
    (e.g., GPT-5 and Gemini~2.5 Pro).%
  }
  \label{fig:accuracy_comparison}
\end{figure*}

Figure~\ref{fig:accuracy_comparison} summarizes the accuracy of all six frontier
LLMs on three reasoning benchmarks under two conditions: direct single-model
inference (\textit{Base LLM}) and our proposed multi-agent framework
(\textit{BIGMAS}).

BIGMAS consistently improves performance across all models and tasks without
exception. The gains are most pronounced for weaker base models: DeepSeek-V3.2
improves from 25.0\% to 36.0\% on Game~24, from 12.0\% to 30.0\% on Six~Fives,
and from 6.0\% to 20.0\% on Tower of London. Similarly, Claude~4.5 Sonnet
improves from 48.0\% to 68.0\%, 15.0\% to 38.0\%, and 40.0\% to 46.0\%
respectively. For already-strong models, BIGMAS still provides meaningful
increments: GPT-5 reaches 100.0\% on both Game~24 and Six~Fives (up from 96.0\%
and 95.0\%), and 98.0\% on Tower of London (up from 91.0\%). Notably, BIGMAS
pushes four out of six models to perfect accuracy on Game~24 (100\%), confirming
that multi-agent coordination saturates performance even on tasks where individual
models already score near-ceiling. 

For Large Reasoning Models (LRMs), the gains are also consistent and substantial.
DeepSeek-V3.2 (+thinking) advances from 92.0\% to 100.0\% on Game~24, from 96.0\%
to 100.0\% on Six~Fives, and from 79.0\% to 87.0\% on Tower of London.
Claude~4.5 (+thinking) shows the most dramatic improvement on Tower of London,
rising from 57.0\% to 93.0\%, suggesting that BIGMAS is particularly effective
at compensating for the planning failures of LRMs on sequential multi-step tasks.
These results establish that multi-agent architectural design provides
\emph{complementary} gains that are orthogonal to model-level reasoning
enhancements via extended chain-of-thought.

\subsection{Comparison with Existing Multi-Agent Frameworks}

To situate BIGMAS within the landscape of existing multi-agent approaches, we
compare it against two representative baselines—ReAct~\cite{yao2022react} and
Tree of Thoughts~\cite{yao2023tree}—using DeepSeek-V3.2 as the backbone model.

\begin{table}[htbp]
\centering
\caption{Accuracy (\%) of DeepSeek-V3.2 under four reasoning frameworks across three tasks.}
\label{tab:method_comparison}
\setlength{\tabcolsep}{8pt}
\begin{tabular}{l ccc}
\toprule
\textbf{Method} & \textbf{Game 24} & \textbf{Six Fives} & \textbf{Tower of London} \\
\midrule
Base LLM               & 25.0 & 12.0 & 6.0 \\
ReAct                  & 26.0 & 18.0 & 10.0 \\
Tree of Thoughts        & 30.0 & 25.0 & 18.0 \\
\textbf{BIGMAS (Ours)} & \textbf{36.0} & \textbf{30.0} & \textbf{20.0} \\
\bottomrule
\end{tabular}
\end{table}

As shown in Table~\ref{tab:method_comparison}, BIGMAS outperforms both
ReAct and Tree of Thoughts across all three tasks. ReAct provides only marginal
gains over the base model (1.0\%--6.0\%), consistent with prior observations that
reactive single-model loops are limited by partial-information decision-making.
Tree of Thoughts improves substantially over ReAct by enabling broader search,
but remains constrained by the absence of a globally shared workspace and dynamic
graph construction. BIGMAS surpasses Tree of Thoughts by 6.0\%, 5.0\%, and 2.0\%
on Game~24, Six~Fives, and Tower of London, respectively, demonstrating that
problem-adaptive graph design combined with centralized workspace coordination
provides a structural advantage over fixed-topology frameworks.

\subsection{Graph Topology Analysis}

\begin{figure}[htbp]
  \centering
  \includegraphics[width=\linewidth]{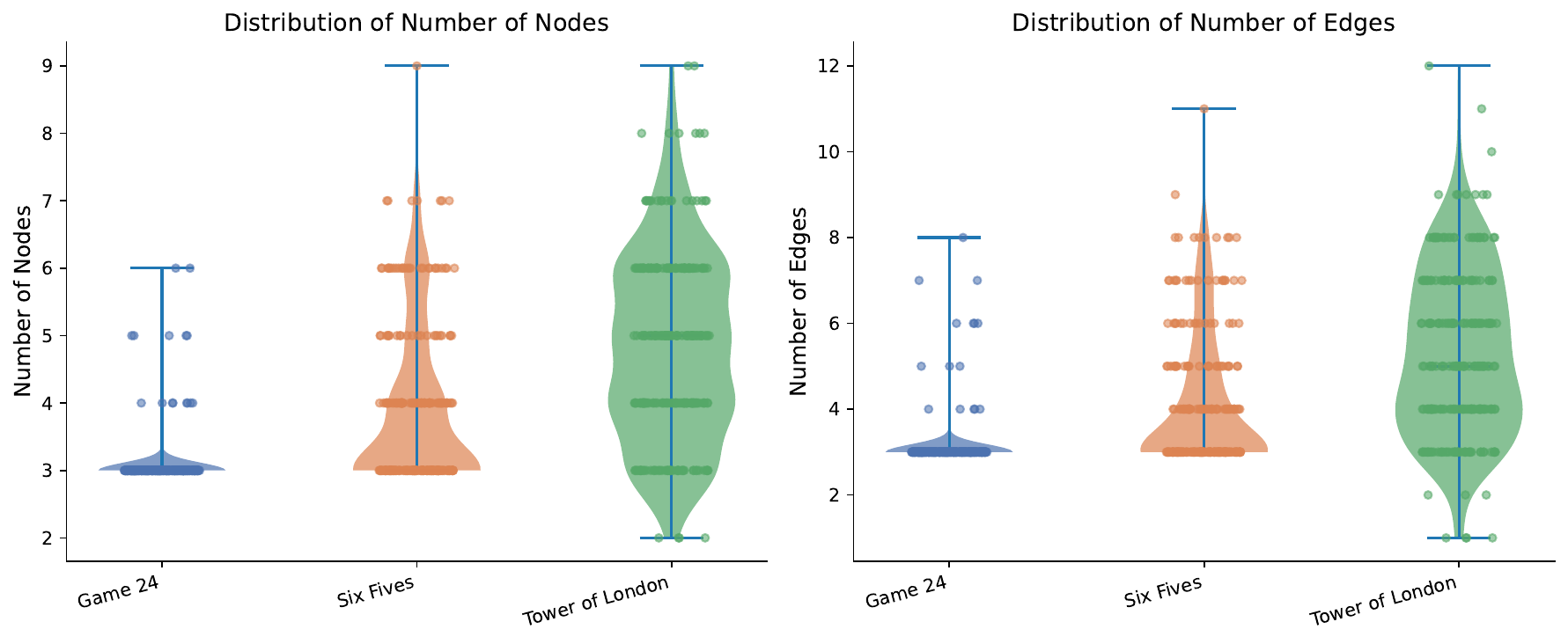}
  \caption{%
    Distribution of graph complexity across three reasoning tasks,
    shown as violin plots overlaid with jittered individual observations.
    The left panel reports the number of nodes and the right panel the
    number of directed edges in each agent graph automatically designed
    by the \textsc{GraphDesigner}.%
  }
  \label{fig:topology_distribution}
\end{figure}

Figure~\ref{fig:topology_distribution} reports the distribution of graph
complexity—measured by node count and directed edge count—across all problem
instances for each task.

The \textsc{GraphDesigner} produces task-appropriate topologies without any
explicit complexity constraints. Game~24 graphs cluster tightly around three
nodes (mean $3.07 \pm 0.38$), reflecting the sufficiency of a compact
\emph{generate}--\emph{validate}--\emph{format} pipeline for arithmetic search.
Six~Fives graphs span three to nine nodes (mean $3.87 \pm 1.12$), with the
higher variance indicating that the model selectively employs richer
expression-search pipelines for harder target values. Tower of London exhibits
the broadest distribution (two to nine nodes, mean $4.82 \pm 1.47$), consistent
with the richer state-space reasoning required for multi-step planning.
Across all tasks, edge counts closely track node counts (mean degree $>1$),
confirming that the designed graphs are relatively dense.

\begin{figure}[htbp]
  \centering
  \includegraphics[width=\linewidth]{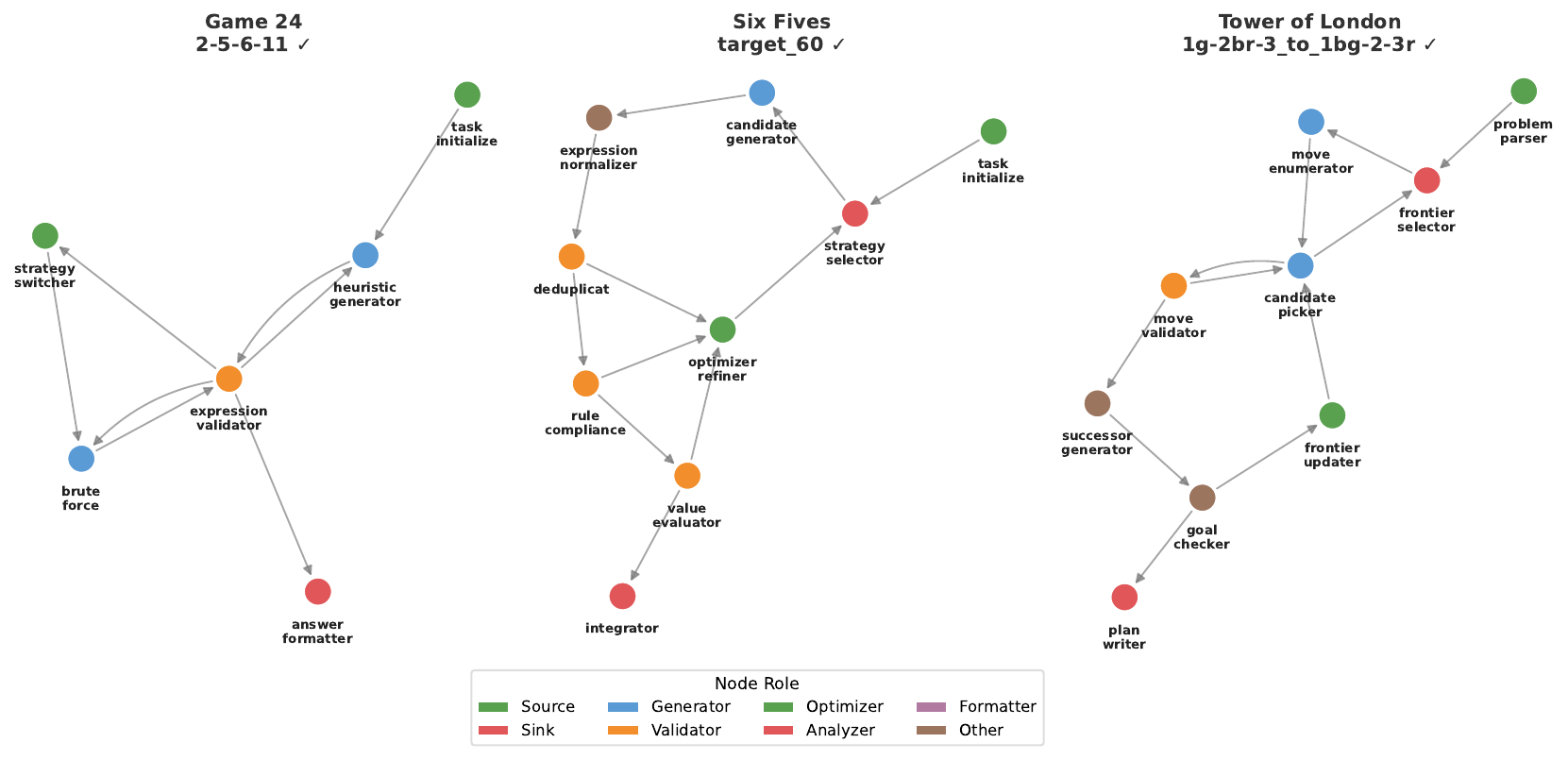}
  \caption{%
Representative agent graph structures automatically designed for
each of the three reasoning tasks.
Each panel displays the highest-node-count graph produced by the
system for that task; all three instances were solved correctly
(indicated by~\checkmark).
}
  \label{fig:representative_graphs}
\end{figure}

Figure~\ref{fig:representative_graphs} presents the highest-node-count
representative graph for each task, all of which were solved correctly. The
Game~24 graph (6 nodes) employs parallel brute-force enumeration and
heuristic generation branches feeding a shared validator. The Six~Fives graph
(9 nodes) chains rule-compliance validation, strategy selection, and value
evaluation into a multi-stage refinement pipeline. The Tower of London graph
(9 nodes) orchestrates BFS-style move enumeration, candidate selection, and
move validation in a cyclic structure suited to state-space search.

\subsection{Node Role Distribution}

\begin{figure}[htbp]
  \centering
  \includegraphics[width=\linewidth]{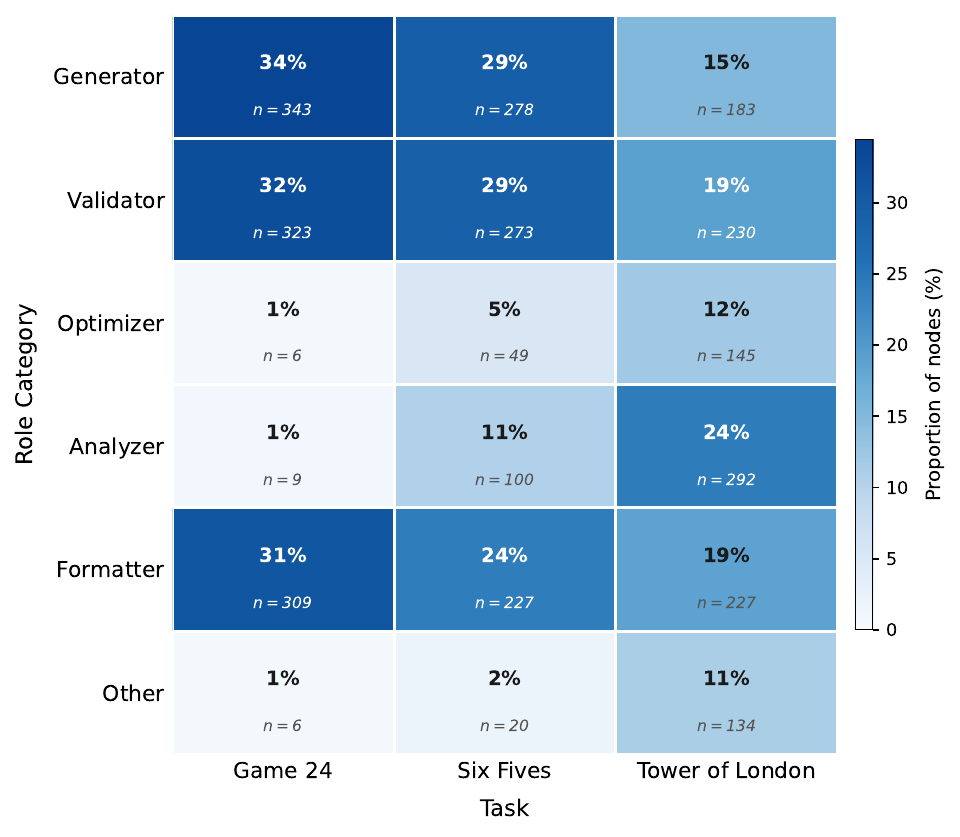}
  \caption{%
    Heatmap of agent node role distribution across three reasoning tasks.
    Each cell reports the proportion (\%) of nodes assigned to a given
    functional role category within all graphs produced for that task;
    raw node counts are shown in italics below each percentage.
    Color intensity encodes the proportion on a white-to-navy scale.%
  }
  \label{fig:node_roles}
\end{figure}

Figure~\ref{fig:node_roles} presents a heatmap of the node role distribution
across tasks, quantifying how \textsc{GraphDesigner} allocates functional
responsibilities.

Role allocations reflect task-specific reasoning requirements. Game~24 graphs
are dominated by Generator (34\%), Validator (32\%), and Formatter (31\%) nodes,
encoding the straightforward \emph{propose}--\emph{verify}--\emph{format}
pattern sufficient for arithmetic search. Six~Fives shows a similar pattern but
with a notable increase in Analyzer (11\%) and Optimizer (5\%) nodes, indicating
that the system introduces expression-search heuristics for the more constrained
generation task. Tower of London exhibits the most diverse role composition:
Analyzer nodes rise to 24\% and Optimizer nodes to 12\%, while Generator and
Formatter proportions decrease substantially, reflecting the rich state-space
reasoning required for multi-step planning. The consistent presence of Formatter
nodes across all tasks ($\geq 19\%$) confirms that the system reliably designates
a dedicated output-formatting stage regardless of task type. Together, these
patterns demonstrate that \textsc{GraphDesigner} implicitly learns task-appropriate
functional decompositions from problem structure alone, without any role
specification in the design prompt.

\subsection{Token Consumption Analysis}

\begin{figure*}[htbp]
  \centering
  \includegraphics[width=\linewidth]{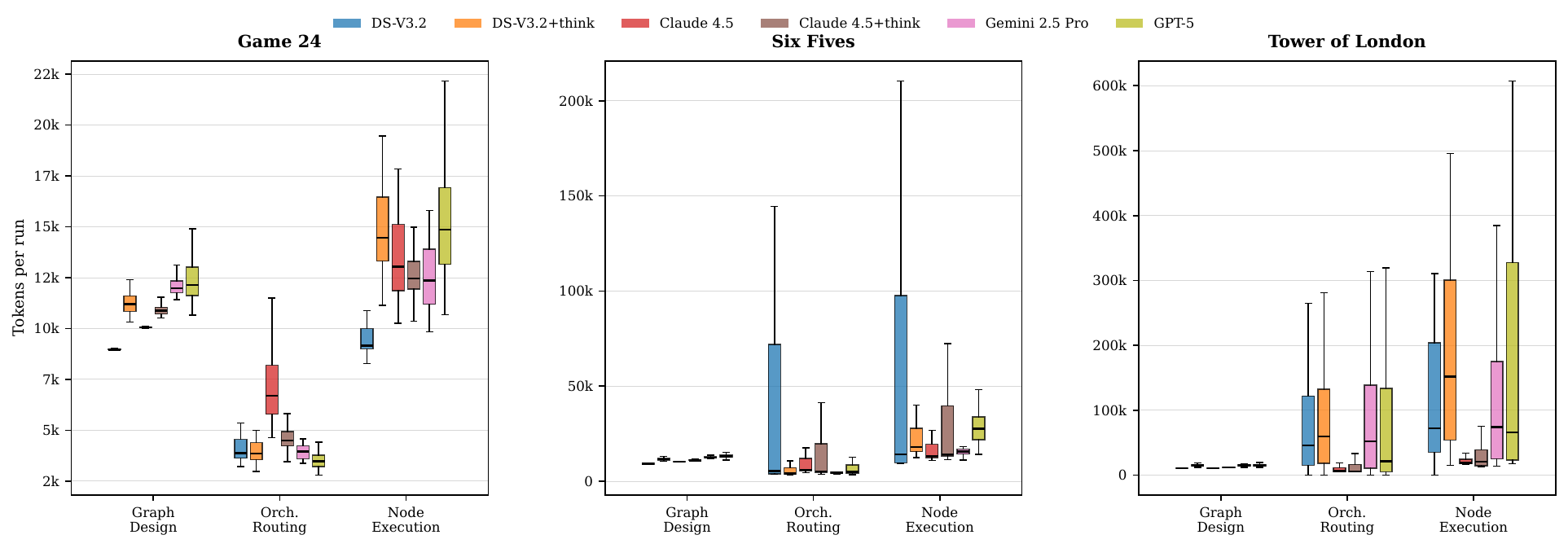}
  \caption{%
    Absolute token consumption per architectural phase across three reasoning
    tasks, shown as grouped box plots (one box per model per phase, outliers
    omitted).
    Node Execution dominates in all tasks, accounting for
    46.4\%, 50.9\%, and 55.5\% of total tokens for
    Game~24, Six~Fives, and Tower of London respectively.
    Orchestrator Routing consumes a bounded but non-trivial share
    (16.9\%--25.2\%), growing with task complexity.
    Graph Design incurs a fixed up-front cost that shrinks as a
    proportion of total tokens when execution steps are numerous.%
  }
  \label{fig:token_phase_abs}
\end{figure*}

Figure~\ref{fig:token_phase_abs} reports token consumption broken down by
architectural phase: Graph Design, Orchestrator Routing, and Node Execution.

Node Execution dominates total token expenditure across all tasks, accounting for
46.4\%, 50.9\%, and 55.5\% of total tokens for Game~24, Six~Fives, and Tower of
London, respectively. This confirms that the majority of computational resources
are allocated to substantive problem-solving rather than system coordination.
Orchestrator Routing consumes a bounded but non-trivial share (16.9\%--25.2\%),
growing with task complexity as broader inter-agent coordination is required.
Graph Design incurs a one-time up-front cost that shrinks as a proportion of
total tokens when execution runs are lengthy (Tower of London: 18.9\% vs.\
Game~24: 36.7\%). The wide interquartile ranges observed for DS-V3.2 and
Gemini~2.5 Pro on Six~Fives and Tower of London reflect high variance in the
number of agent-loop iterations triggered for harder instances, consistent with
the routing count analysis below.

\subsection{Orchestrator Routing Behavior}

\begin{figure}[htbp]
  \centering
  \includegraphics[width=\linewidth]{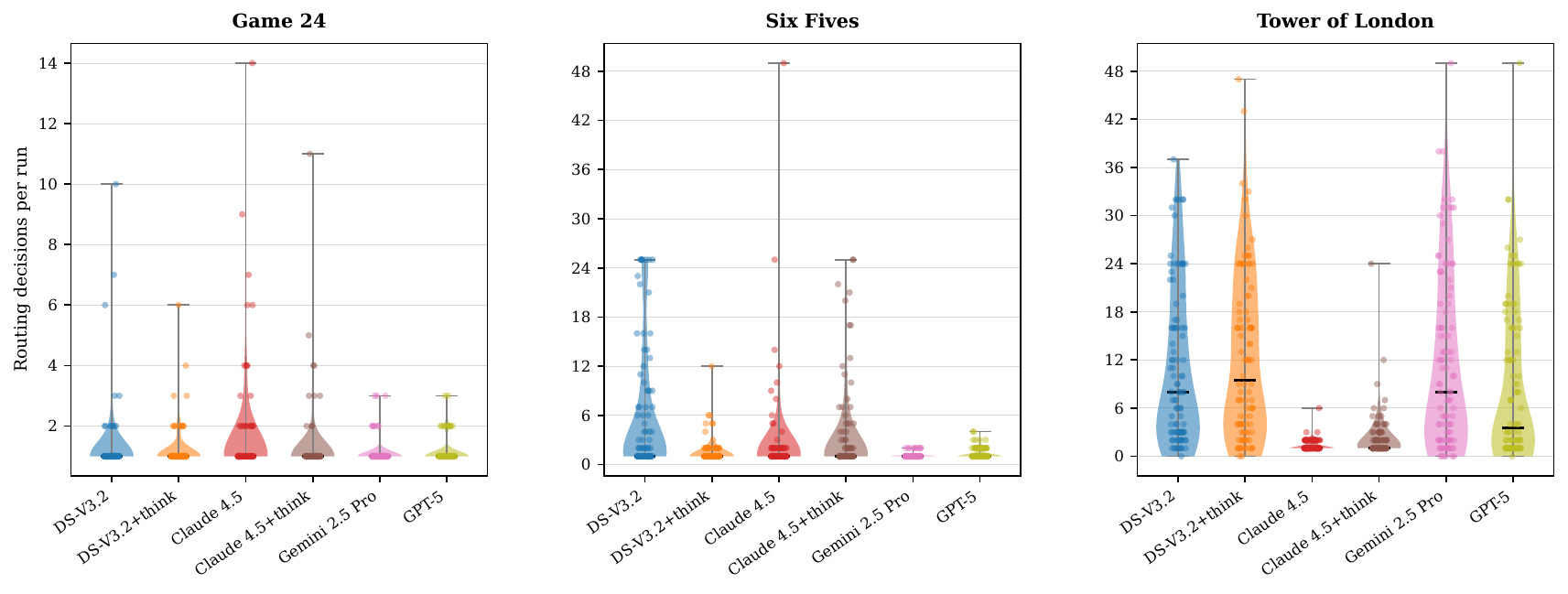}
  \caption{%
    Distribution of orchestrator routing decisions per run across three
    reasoning tasks, displayed as violin plots overlaid with jittered
    individual observations.
    Routing count serves as a proxy for execution-time complexity.
    Game~24 concentrates near one decision per run (mean $1.3$),
    Six~Fives shows moderate spread (mean $2.6$), and Tower of London
    yields the highest and most variable counts (mean $7.9$, maximum $49$).%
  }
  \label{fig:routing_count_dist}
\end{figure}

Figure~\ref{fig:routing_count_dist} presents the distribution of orchestrator
routing decisions per run, serving as a proxy for execution-time complexity.

Routing counts scale naturally with task difficulty without any hard-coded
scheduling. Game~24 concentrates near one decision per run (mean $1.3$),
consistent with its compact three-node pipelines. Six~Fives shows moderate
spread (mean $2.6$), with DS-V3.2 as a notable outlier (mean $5.7$) due to
its tendency to design larger iterative graphs. Tower of London yields the
highest and most variable counts (mean $7.9$, maximum 49), reflecting the
iterative state-space search required for multi-step planning.

\begin{figure}[htbp]
  \centering
  \includegraphics[width=\linewidth]{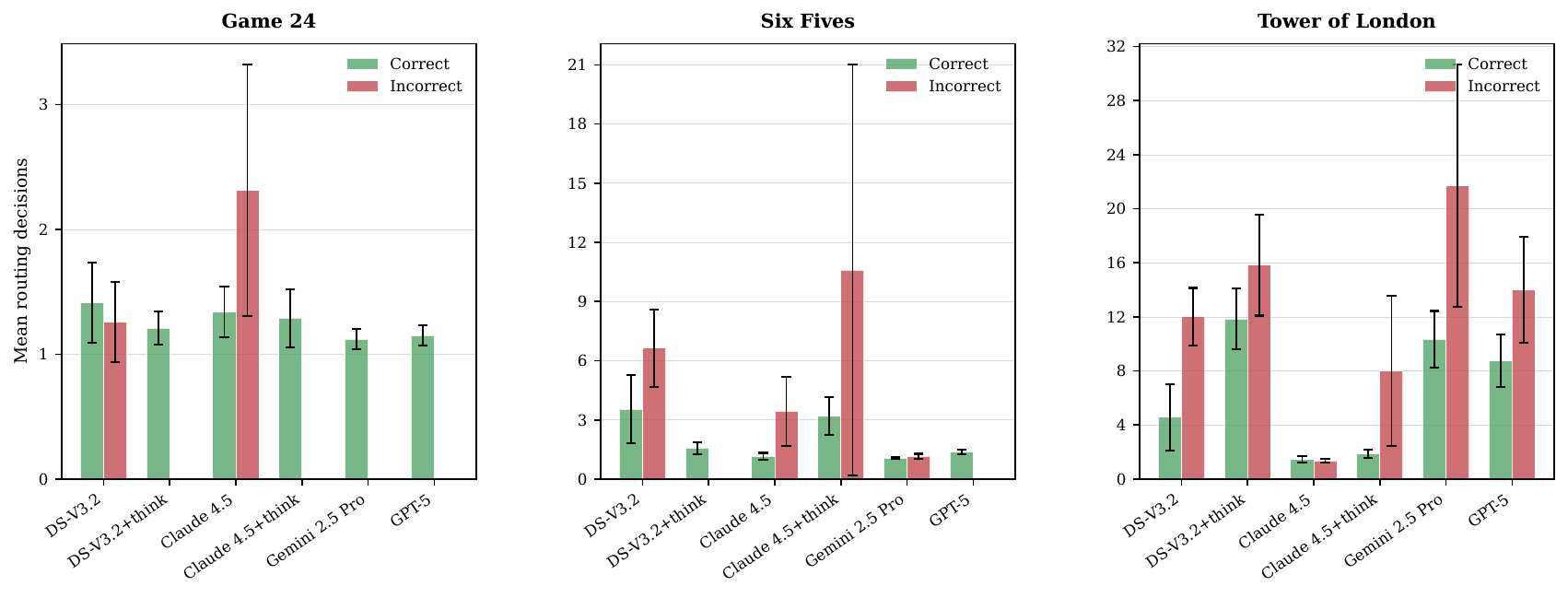}
  \caption{%
    Mean number of orchestrator routing decisions for correctly solved
    (green bars) versus incorrectly solved (red bars) instances,
    broken down by model and task.
    Error bars denote 95\% confidence intervals.
    Incorrect runs consistently require more routing calls than correct ones
    across all three tasks.%
  }
  \label{fig:routing_vs_accuracy}
\end{figure}

Figure~\ref{fig:routing_vs_accuracy} further reveals a characteristic failure
mode: incorrect runs consistently require \emph{more} routing decisions than
correct ones.

The gap between failed and successful runs is modest for Game~24
($1.2$ vs.\ $1.6$ decisions on average), pronounced for Six~Fives
($1.9$ vs.\ $4.5$), and substantial for Tower of London ($7.3$ vs.\ $9.4$).
This pattern exposes a characteristic failure mode: when an agent graph fails
to find a valid solution, the Orchestrator continues cycling through nodes
until the step budget is exhausted, accumulating routing calls without
converging. Conversely, successful runs terminate earlier because the
Orchestrator correctly detects a validated solution and routes to the sink
node. This finding suggests that routing-count thresholds could serve as
lightweight early-stopping signals to curtail unproductive execution on hard
instances, a direction we leave for future work.

\section{Discussion and Conclusion}
\label{sec:discussion}
 
This work introduced BIGMAS, a brain-inspired graph multi-agent framework grounded in Global Workspace Theory, in which specialized LLM agents are organized as nodes in a dynamically constructed directed graph and coordinate exclusively through a centralized shared workspace. The framework instantiates three principles drawn from flexible human cognition: processor specialization through role-specific agent nodes, dynamic coalition formation through per-problem graph construction by a GraphDesigner agent, and global broadcast through a workspace that maintains a consistent view of task state across all agents.
 
A central finding is that multi-agent architectural coordination provides gains that are \emph{orthogonal} to model-level reasoning capability. Prior systematic investigations have shown that LRMs suffer accuracy collapse beyond task-specific complexity thresholds\cite{shojaee2025illusion}, and that providing explicit solution algorithms does not alleviate this collapse---pointing to consistent logical execution, not solution discovery, as the fundamental bottleneck. Our results directly support this diagnosis: BIGMAS improves performance across all six frontier models tested, including LRMs already equipped with self-reflection and extended reasoning traces. The most striking instance is Claude~4.5 (+thinking) on Tower of London, where BIGMAS delivers a 36-percentage-point gain despite the model already performing extended internal reasoning. This indicates the benefit of BIGMAS is structural: by decomposing a problem into specialized subagents coordinated through a global workspace, the system reduces the per-agent cognitive burden and externalizes intermediate state in a form that is globally visible, writable, and verifiable---a mechanism that is unavailable to any single-model approach regardless of reasoning depth.
 
The graph topology and node role analyses provide empirical support for these design principles. The GraphDesigner does not produce a fixed pipeline applied uniformly across problems; rather, it constructs structurally distinct graphs whose complexity tracks task demands---from compact three-node generate--validate--format pipelines for Game~24 to nine-node cyclic structures for Tower of London. This mirrors the brain's dynamic large-scale network reconfiguration\cite{bassett2011dynamic,finc2020dynamic}: routine tasks engage relatively isolated modules, while complex problems recruit broader functional coalitions. The autonomous emergence of task-appropriate role distributions---with Analyzer and Optimizer nodes rising substantially from arithmetic to planning tasks, without any task-specific engineering---further suggests that the GraphDesigner is sensitive to the underlying reasoning demands of each problem class in a way that parallels prefrontal orchestration of specialized cortical areas. This parallels findings in computational neuroscience showing that task-appropriate functional modules emerge spontaneously from connectivity constraints rather than explicit programming~\cite{minsky1986society}, and is consistent with hierarchical society-of-mind principles in which specialized sub-agents cooperate without centralized role assignment.
 
The orchestrator routing analysis reveals an additional emergent property: routing count functions as a natural proxy for instance-level difficulty, arising without explicit scheduling logic purely from global conditioning on workspace state and execution history. The systematic divergence between correct and incorrect runs---most pronounced on Tower of London (7.3 vs.\ 9.4 routing decisions)---exposes a characteristic failure mode in which the orchestrator continues cycling when no valid solution has been found. This connects to broader findings on LLM self-evaluation and the difficulty of knowing when to stop in iterative refinement\cite{zeng2025evolving,han2025stitch,madaan2023selfrefine}: global workspace visibility is necessary but not sufficient for reliable termination, and detecting non-convergence remains an open problem. The token consumption analysis complements this picture: Node Execution accounts for 46.4\%--55.5\% of total tokens across tasks, confirming that coordination overhead remains bounded and that BIGMAS is most token-efficient precisely on the hard tasks---Tower of London at 18.9\% Graph Design overhead---where it is most needed.
 
Compared to existing multi-agent frameworks, BIGMAS advances the state of the art along two dimensions simultaneously. Reactive approaches such as ReAct and Reflexion operate with partial information and fixed decision loops; planning-execution frameworks\cite{erdogan2025plan} such as LLMCompiler\cite{kim2024llm} introduce upfront global planning but use static, pre-specified topologies. BIGMAS combines the benefits of both: per-problem graph construction provides the structural flexibility that static frameworks lack, while global workspace coordination provides the full-state visibility that reactive frameworks lack. The gains over Tree of Thoughts---which already incorporates broad search---further confirm that the advantage of BIGMAS does not reduce to more extensive exploration, but stems from the synergy between adaptive topology and shared state.
 
Taken together, these results establish that the architectural organization of reasoning systems---not merely the capacity or reasoning strategy of individual model components---is a fundamental determinant of performance on complex tasks. Multi-agent coordination and model-level scaling are complementary rather than competing investments: the former provides a structural remedy for the accuracy collapse that the latter cannot resolve alone. As frontier models continue to improve, BIGMAS-style architectures offer a principled, neuroscience-grounded path for extracting further gains from those models on the hardest reasoning problems---a path that becomes more, not less, valuable as individual model ceilings are approached.

\section{Limitations and Future Directions}
\label{sec:limitations}
 
The current evaluation is limited to three combinatorial reasoning benchmarks; extending BIGMAS to open-domain question answering, mathematical competition problems, and code generation remains important future work. The GraphDesigner operates without memory of prior designs, treating each instance independently; incorporating episodic memory or meta-learned graph construction could improve both efficiency and quality across recurring problem families. The step budget $T_{\max}$ and self-correction limit $R$ are fixed hyperparameters, though the routing dynamics analysis suggests that routing-count signals could support adaptive early stopping to curtail non-converging runs. Finally, while per-agent reasoning burden is reduced through decomposition, total token cost exceeds single-model inference; token-aware graph design and learned role specialization via fine-tuned specialist agents represent promising directions toward more efficient and capable heterogeneous multi-agent systems.

\bibliographystyle{IEEEtran}
\bibliography{ref}

\vfill

\end{document}